%% file: ms2sl-acl.tex
\pdfoutput=1

\documentclass[11pt]{article}

\usepackage[]{acl}

\usepackage{times}
\usepackage{latexsym}

\usepackage[T1]{fontenc}

\usepackage[utf8]{inputenc}

\usepackage{microtype}

\usepackage{inconsolata}

\usepackage[utf8]{inputenc} 
\usepackage[T1]{fontenc}    
\usepackage{url}            
\usepackage{booktabs}       
\usepackage{amsfonts}       
\usepackage{nicefrac}       
\usepackage{microtype}      
\usepackage{xcolor}         
\usepackage{bbding}         
\usepackage{graphicx}
\usepackage{tabularx}       
\usepackage{comment}
\usepackage{color}

\usepackage{amsmath}
\usepackage{amssymb}
\usepackage{float}
\usepackage[symbol]{footmisc}
\usepackage{multirow}
\usepackage{bm}
\usepackage{arydshln}
\usepackage{subcaption}
\usepackage{makecell}
\usepackage{enumitem}
\usepackage{wrapfig}
\usepackage{authblk}

\usepackage{CJKutf8}

\DeclareRobustCommand\onedot{.}

\definecolor{mygray}{gray}{.9}
\definecolor{ggray}{RGB}{127,127,127}
\definecolor{myred}{RGB}{192,0,0}
\definecolor{redb}{RGB}{217,148,143}
\definecolor{myyellow}{RGB}{190,144,0}
\definecolor{mygreen}{RGB}{80,100,40}
\definecolor{myblue}{RGB}{30,90,100}

\def\eg{\textit{e.g}\onedot} 
\def\ie{\textit{i.e}\onedot}

\makeatletter
\newcommand{\thickhline}{
    \noalign {\ifnum 0=`}\fi \hrule height 1pt
    \futurelet \reserved@a \@xhline
}
\newcommand{\et}[2]{${#1}^{\pm{#2}}$}
\newcommand{\etr}[2]{$\textcolor{red}{{#1}}^{\pm{#2}}$}
\newcommand{\etbb}[2]{$\textcolor{blue}{{#1}}^{\pm{#2}}$}

\makeatletter
            
\usepackage[capitalize]{cleveref}
\crefname{section}{Sec.}{Secs.}
\Crefname{section}{Section}{Sections}
\Crefname{table}{Table}{Tables}
\crefname{table}{Tab.}{Tabs.}

\input{macros.tex}

\title{MS2SL: Multimodal Spoken \\ Data-Driven  Continuous Sign Language Production}

\author[1,3]{\textbf{Jian Ma}}
\author[2]{\textbf{Wenguan Wang}}
\author[2]{\textbf{Yi Yang}}
\author[1]{\textbf{Feng Zheng}$^\dag$}
\affil[1]{Southern University of Science and Technology}
\affil[ ]{\textsuperscript{2}ReLER, CCAI, Zhejiang University \hspace{3mm} \textsuperscript{3}ReLER, University of Technology Sydney}
\affil[ ]{\href{https://hechang25.github.io/MS2SL}{\textcolor[rgb]{0.50,0.00,0.25}{https://hechang25.github.io/MS2SL}}}

\begin{document}
\maketitle
\begingroup\def\thefootnote{\dag}\footnotetext{\small Corresponding author.}\endgroup
\begin{abstract}
Sign language understanding has made significant strides; however, there is still no viable solution for generating sign sequences directly from entire spoken content, \eg,~text or speech.
In this paper, we propose a unified framework for continuous sign language production, easing communication between sign and non-sign language users. 
In particular, a sequence diffusion model, utilizing embeddings extracted from text or speech, 
is crafted to generate sign predictions step by step.
Moreover, by creating a joint embedding space for text, audio, and sign, 
we bind these modalities and leverage the semantic consistency among them to provide informative feedback for the model training.
This embedding-consistency learning strategy minimizes the reliance on sign triplets and ensures continuous model refinement, even with a missing audio modality.
Experiments on How2Sign and PHOENIX14T datasets demonstrate that our model achieves competitive performance in sign language production.
\end{abstract}

\input{sections/intro.tex}
\input{sections/related.tex}
\input{sections/method.tex}
\input{sections/experiment.tex}
\input{sections/conclusion.tex}
\input{sections/Limitations.tex}
\bibliographystyle{acl_natbib}
\bibliography{custom}
\input{sections/appendix.tex}
\end{document}

%% file: macros.tex
\usepackage{hyphenat}
\usepackage{soul,color}
\usepackage{amsopn}

\definecolor{Red}{cmyk}{0,1,1,0}
\definecolor{Green}{cmyk}{1,0,1,0}
\definecolor{Cyan}{cmyk}{1,0,0,0}
\definecolor{Purple}{cmyk}{0.45,0.86,0,0}
\definecolor{Rosolic}{cmyk}{0.00,1.00,0.50,0}
\definecolor{Blue}{cmyk}{1.00,1.00,0.00,0}
\definecolor{BlueViolet}{cmyk}{0.86,0.91,0,0.04}
\definecolor{NavyBlue}{cmyk}{0.94,0.54,0,0}

\newcommand{\hidden}[1]{{\color{NavyBlue}}}



%% file: sections/intro.tex
\section{Introduction}
\label{intro} 
\vspace{-5pt}
Sign$_{\!}$ language,$_{\!}$ a$_{\!}$ visual$_{\!}$ language,$_{\!}$ combines$_{\!}$ both$_{\!}$ manual$_{\!}$ (hand$_{\!}$ gestures)$_{\!}$ and$_{\!}$ non-manual$_{\!}$ cues$_{\!}$ for$_{\!}$ communication.$_{\!}$ It$_{\!}$ is$_{\!}$ specifically$_{\!}$ designed$_{\!}$ for$_{\!}$ the deaf$_{\!}$ and$_{\!}$ hearing-impaired community$_{\!}$~\citep{hickok1996neurobiology,armstrong2003origins,campbell2008sign,DBLP:conf/aaai/ZhouZZL20}.$_{\!}$ According to the World Federation of the Deaf, there are $70$ million deaf people and more than $200$ kinds of sign languages in the world~\cite{fenlon2015sign,DBLP:journals/eswa/Nunez-MarcosPL23}.
Improvements in sign language production~(SLP) can bridge the communication gap between the deaf$_{\!}$ and$_{\!}$ hearing$_{\!}$~\cite{mehdi2002sign,harris2009research,DBLP:conf/tsp/TaskiranKK18,DBLP:journals/eswa/RastgooKE21,DBLP:journals/uais/KahlonS23,LuoY24}.

The challenges primarily arise from \textbf{phonological difference} and \textbf{data scarcity}. Phonological$_{\!}$ difference:$_{\!}$ signs$_{\!}$ are$_{\!}$ composed$_{\!}$ of$_{\!}$ various$_{\!}$ manual$_{\!}$ and$_{\!}$ non-manual features~\cite{mann2010acquisition},$_{\!}$ such$_{\!}$ as hand gestures, facial expressions and limb movements~\cite{liddell1989american,johnson2011toward,sandler2012phonological}.
The differences in phonological structure and means of expression create challenges in modeling the two languages. 
Data scarcity: multimodal high-quality sign language datasets are relatively scarce, and some datasets tend to be specific to a particular language or domain,~\eg,~American sign~\cite{DBLP:conf/cvpr/DuartePVGDMTG21}, German weather~\cite{DBLP:conf/lrec/ForsterSKBN14,DBLP:conf/cvpr/CamgozHKNB18}.
Furthermore, hearing impairments hinder pronunciation~\cite{moeller2000early,yoshinaga2003screening}, making it strenuous to collect sign video with 
aligned audio and usually resulting in the lack of auditory information. 
Previous$_{\!}$ researches~\cite{DBLP:conf/icmcs/ZhangZXPL16,DBLP:conf/iccv/CamgozHKB17,DBLP:conf/aaai/HuZL21,DBLP:conf/iccv/HuZZWL21,DBLP:conf/iciai/YinTLH22} primarily focused on sign language recognition, which identifies sign fragments as the corresponding sign language lexicons~(\eg, gloss). 
Several work~\cite{DBLP:conf/eccv/SaundersCB20,DBLP:journals/ijcv/SaundersCB21,DBLP:conf/cvpr/SaundersCB22,DBLP:conf/bmvc/HwangKP21,DBLP:journals/corr/abs-2210-06312} 
$_{\!}$ manage$_{\!}$ the$_{\!}$ transition$_{\!}$ from$_{\!}$ gloss$_{\!}$ to$_{\!}$ sign$_{\!}$ sequences,$_{\!}$ yet the grammar of gloss can be perplexing for those without sign language training.$_{\!}$
\begin{figure}[tp]
    \vspace{-8pt}
    \centering
    \includegraphics[width=0.47\textwidth, trim=19bp 10bp 20bp 10bp, clip]{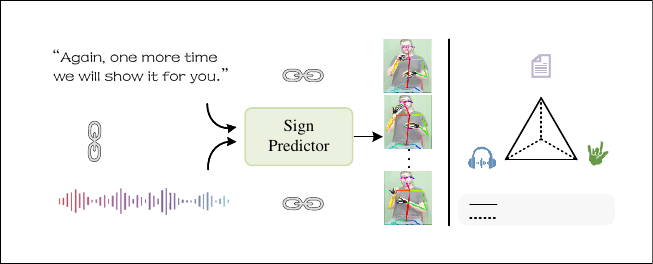}
    \put(-190,36){\scalebox{.55}{Binding}}
    \put(-125,68){\scalebox{.55}{Binding}}
    \put(-125,20){\scalebox{.55}{Binding}}
    \put(-39,13){\scalebox{.5}{Coexistent}}
    \put(-39,7){\scalebox{.5}{Simultaneous}}
    \put(-32,56){\scalebox{.58}{Text}}
    \put(-57,22){\scalebox{.58}{Audio}}
    \put(-10,22){\scalebox{.58}{Sign}}
    \vspace{-10pt}
    \caption{\textbf{Illustration of our sign language producer.} 1) We propose a unified, multimodal spoken data-driven framework for SLP that can directly produce sign sequences from spoken text or speech audio. 2) To overcome data scarcity, we train a joint embedding space through the spontaneous alignment of multimodal data. Within this space, we establish a consistency learning strategy to provide feedback signals that boost training. }
    \label{fig:TE}
    \vspace{-22pt}
\end{figure}
\citet{DBLP:conf/eccv/SaundersCB20, SaundersCB21} can transcribe discrete words or phrases into continuous sign language sequences.
However, directly producing continuous signs from entire spoken sentences still remains more exploration and efforts.

To promote barrier-free communication between signers and speakers, we introduce a \underline{M}ultimodal \underline{S}poken Data-Driven Continuous \underline{S}ign \underline{L}anguage Production~(MS2SL) framework~(Fig.~\ref{fig:TE}).
MS2SL can animate sign keypoint sequences from either$_{\!}$ speech$_{\!}$ audio$_{\!}$ or$_{\!}$ text.$_{\!}$
In$_{\!}$ addition,$_{\!}$ to$_{\!}$ alleviate$_{\!}$ data$_{\!}$ demands,$_{\!}$ we$_{\!}$ adopt$_{\!}$ an embedding-consistency learning~(ECL) strategy, which is inherently based on the reciprocity among modalities, to bolster the model training.
Specifically, MS2SL initially employs pre-training models like CLIP~(text)~\cite{DBLP:conf/icml/RadfordKHRGASAM21} and HuBERT~(audio)~\cite{DBLP:journals/taslp/HsuBTLSM21} to extract features from input.
Subsequently,$_{\!}$ we$_{\!}$ utilize$_{\!}$ these$_{\!}$ features,$_{\!}$ serving as control conditions for the diffusion, to generate sign sequences.
The attention mechanism~\cite{DBLP:conf/nips/VaswaniSPUJGKP17} is employed to model the relationships among conditions, denoising steps, and sign movements. 
Besides$_{\!}$ that,$_{\!}$ ECL$_{\!}$ does$_{\!}$ not$_{\!}$ require$_{\!}$ the three modalities to coexist in the dataset.
By learning a joint embedding space, inspired by ImageBind~\cite{DBLP:conf/cvpr/GirdharELSAJM23}, ECL tightly binds the properties of different modalities and generates feedback signals to boost the training process.
First, we utilize contrastive learning to bind audio and text in the embedding space.
Then, we leverage the semantic consistency between co-occurring data to infer and reconstruct the embedding of missing modalitiy.
The reconstruction error between the generated signs and groundtruth can be used to iteratively update MS2SL until convergence.
ECL can foster cross-learning between different generation streams, allowing training even in the absence of certain modality.
Furthermore, the inclusion of audio data not only enriches sample diversity and enhances multimodal comprehension but also assists in accurately capturing the expression and semantic content of sign language.
We validate the effectiveness of our method across two prevalent datasets How2Sign~\cite{DBLP:conf/cvpr/DuartePVGDMTG21} and  PHOENIX14T~\cite{DBLP:conf/cvpr/CamgozHKNB18}. Experimental results demonstrate that MS2SL achieves SOTA performance, both in terms of semantic consistency and sign accuracy.
In conclusion, our primary contributions are outlined as follows:
\begin{itemize} [leftmargin=*]
    \vspace{-10pt}
    \item We propose MS2SL, a unified diffusion framework for efficient multimodal spoken to sign language production. 
        MS2SL is able to directly convert entire speech or text sentences into corresponding sign keypoints sequences.
    \vspace{-10pt}
    \item We present an ECL strategy that leverages the intrinsic relations to enhance data utilization.
    \vspace{-10pt}
    \item We show that joint embedding is suitable for generative tasks that are prone to modality missing.
\end{itemize} 

%% file: sections/related.tex
\section{Related Work}
\vspace{-6pt}
\label{relatedwork} \noindent\textbf{Sign Language Understanding.} Similar to spoken language, sign language follows specific linguistic rules~\cite{sandler2006sign,brentari2011sign,petitto2016visual,sandler2017challenge}.
Existing researches are primarily dedicated to sign language translation~(SLT) and recognition.
SLT typically involves translating sign language into spoken language~\cite{DBLP:conf/cvpr/CamgozHKNB18,DBLP:journals/corr/abs-2202-03086,DBLP:conf/cvpr/CamgozKHB20,LinWZSZY23}.
Sign language recognition~\cite{DBLP:journals/tmm/AdaloglouCPSPZX22,SelvarajNKK22} means interpreting and classifying of body movements in videos, covering isolated~\cite{DBLP:conf/conll/ImashevMKS20} and continuous signs~\cite{DBLP:conf/cvpr/CuiLZ17,DBLP:conf/cvpr/CamgozHKNB18,DBLP:conf/cvpr/CamgozKHB20}.
SLP~\cite{DBLP:conf/cvpr/Shalev-Arkushin23} is the process of creating sign sequences from spoken text, and can be seen as the reverse process of SLT.
These existing studies on SLT and SLP primarily focus on converting between sign videos and discrete glosses, either directly or indirectly.
A few of Text2Sign works~\cite{DBLP:conf/eccv/SaundersCB20,DBLP:journals/ijcv/SaundersCB21,SaundersCB21} are grounded in datasets with relatively homogeneous  scenario~\cite{DBLP:conf/cvpr/CamgozHKNB18} and discrete spoken transcriptions.

\noindent\textbf{Diffusion Model.} The diffusion model demonstrates exceptional proficiency in various generative tasks~\cite{DBLP:conf/nips/HoJA20,DBLP:conf/iccv/ChoiKJGY21,DBLP:conf/cvpr/LugmayrDRYTG22,DBLP:conf/cvpr/AvrahamiLF22}. 
Beyond image generation, diffusion models also perform well in generating sequence data~\cite{DBLP:journals/corr/abs-2212-10325,DBLP:journals/corr/abs-2305-09515}.
In recent years, some work has begun to apply diffusion models to SLP.
By iteratively updating information, diffusion models can gradually infer the distribution of subsequent data, thereby providing more accurate and coherent results.
Ham2Pose~\cite{DBLP:conf/cvpr/Shalev-Arkushin23} leverages diffusion to animate HamNoSys, a lexicon of sign symbols, into sign keypoint sequences.
Though impressive, Ham2Pose can only produce videos with a single sign symbol, falling short in conveying sentences with complete semantics.

\noindent\textbf{Cross-modal Consistency Learning.} Deep learning often requires ample labeled data to work properly. However, the cost of collecting sign data is prohibitive and audio data is often lacking. Recent methods enhance model training by applying consistency training to massive unlabeled data~\cite{DBLP:conf/nips/BachmanAP14,DBLP:conf/nips/SajjadiJT16,DBLP:conf/emnlp/ClarkLML18,DBLP:journals/pami/MiyatoMKI19}.
The principle of consistency learning, employing the cyclical duality between different tasks or data as feedback signals to regularize training~\cite{DBLP:conf/nips/HeXQWYLM16}, has its roots in the domain of language translation~\cite{DBLP:conf/iccv/YiZTG17,DBLP:conf/nips/LuKYPB17,DBLP:conf/acml/ZhaoXQXL20}.
It primarily encompasses inter-task (dual-learning) and intra-task (cycle-consistency learning) varieties.
Dual-learning simultaneously trains bidirectional mapping functions between tasks, creating a primal-dual pair where one function's output approximates the input of the inverse function~\cite{DBLP:conf/iccv/YiZTG17,DBLP:conf/cvpr/WangLSGW22,DBLP:conf/cvpr/ZhangXHL18,DBLP:conf/cvpr/ShahCRP19,DBLP:conf/iclr/WangXHTQZL19,DBLP:conf/acml/ZhaoXQXL20,DBLP:conf/nips/XieDHL020}.
Cycle-consistency learning is designed to enhance the self-reconstruction capabilities of samples produced intrinsically by the same model~\cite{DBLP:conf/iccv/ZhuPIE17,DBLP:conf/icml/AlmahairiRSBC18,DBLP:conf/cvpr/RaoHILIK20,DBLP:conf/cvpr/MathewNKK20}.
However, these methods frequently emphasize the duality between two tasks or modalities, overlooking the interplay and mutual influence among multimodal data within the same task.

Limited studies focus on directly generating sign language sequences from entire spoken sentences. To our best knowledge, we are the pioneers in effecting this conversion.
This study harnesses sequential diffusion models to incrementally generate noise predictions, enabling cross-modal sign language generation.$_{\!}$
With$_{\!}$ the$_{\!}$ help$_{\!}$ of$_{\!}$ ECL,$_{\!}$ MS2SL$_{\!}$ can generate various feedback signals even in the absence of co-occurring ternary data: assessing the reconstruction loss with the signs generated from the reconstructed audio embeddings.
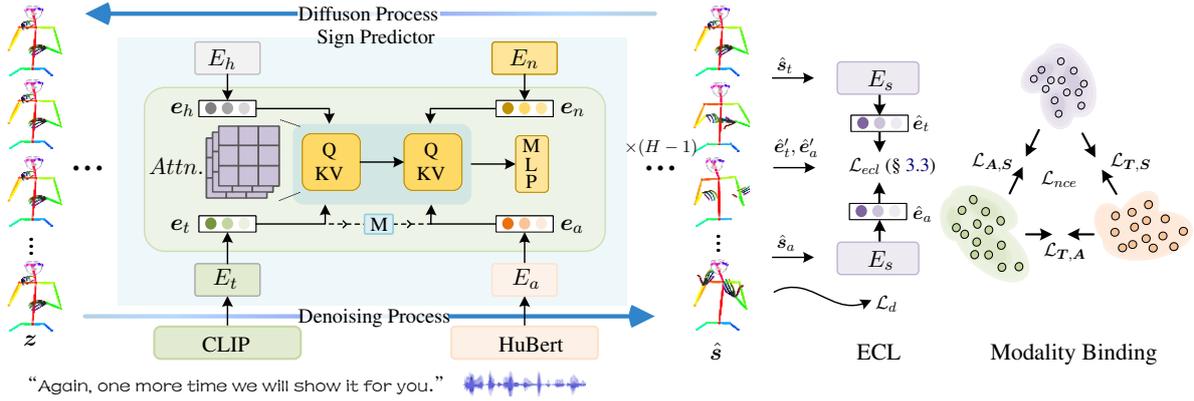
\begin{figure*}[ht]
  \vspace{-8pt}
    \includegraphics[width=\linewidth, trim=20bp 20bp 10bp 20bp, clip]{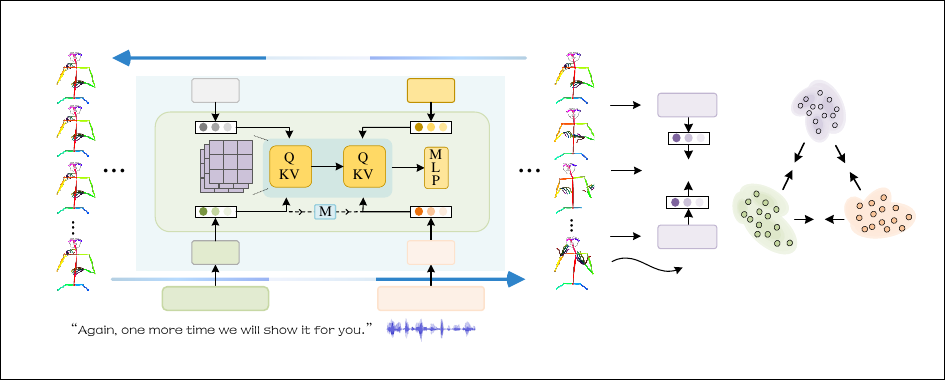}
    \put(-260,125){\scalebox{.80}{$E_n$}}
    \put(-259,42){\scalebox{.80}{$E_a$}}
    \put(-373,125){\scalebox{.80}{$E_{h}$}}
    \put(-371,42){\scalebox{.80}{$E_t$}}
    \put(-126,117){\scalebox{.80}{$E_s$}}
    \put(-126,49){\scalebox{.80}{$E_s$}}
    \put(-185,14){\scalebox{.80}{$\hat{\bm{s}}$}}
    \put(-442,19){\scalebox{.80}{$\bm{z}$}}
    \put(-339,142){\scalebox{.70}{Diffuson Process}}
    \put(-332,133){\scalebox{.70}{Sign Predictor}}
    \put(-339,28){\scalebox{.70}{Denoising Process}}
    \put(-375,17){\scalebox{.75}{CLIP}}
    \put(-264,17){\scalebox{.75}{HuBert}}
    \put(-241,107){\scalebox{.80}{$\bm{e}_n$}}
    \put(-241,63){\scalebox{.80}{$\bm{e}_a$}}
    \put(-387,107){\scalebox{.80}{$\bm{e}_{h}$}}
    \put(-387,63){\scalebox{.80}{$\bm{e}_t$}}
    \put(-395,84){\scalebox{.80}{$Attn.$}}
    \put(-217,92){\scalebox{.60}{$\times(H-1)$}}
    \put(-35,87){\scalebox{.65}{$\mathcal{L}_{\bm{T}, \bm{S}}$}}
    \put(-87,87){\scalebox{.65}{$\mathcal{L}_{\bm{A}, \bm{S}}$}}
    \put(-60,53){\scalebox{.65}{$\mathcal{L}_{\bm{T}, \bm{A}}$}}
    \put(-133,85){\scalebox{.65}{$\mathcal{L}_{ecl}$~(\S~\ref{sec:ECL})}}
    \put(-61,80){\scalebox{.65}{$\mathcal{L}_{nce}$}}
    \put(-123,33){\scalebox{.65}{$\mathcal{L}_{d}$}}
    \put(-161,92){\scalebox{.65}{$\hat{\bm{e}}'_t, \hat{\bm{e}}'_a$}}
    \put(-130,14){\scalebox{.80}{ECL}}
    \put(-80,14){\scalebox{.80}{Modality Binding}}
    \put(-160,123){\scalebox{.65}{$\hat{\bm{s}}_t$}}
    \put(-160,56){\scalebox{.65}{$\hat{\bm{s}}_a$}}
    \put(-109,101){\scalebox{.65}{$\hat{\bm{e}}_t$}}
    \put(-109,67){\scalebox{.65}{$\hat{\bm{e}}_a$}}
    
  \vspace{-2pt}
  \caption{\textbf{Overview of our framework for MS2SL.} It includes three key components: sign predictor~(\S\ref{sec:SP}), modality binding~(\S\ref{sec:MB}) and ECL strategy~(\S\ref{sec:ECL}). MS2SL directly unifies spoken content from different modalities into a common sign language production framework. The introduction of the joint embedding space and ECL reduces the reliance on co-occurring (text, audio, sign)~triplet.}
  \label{fig:framework}
  \vspace{-12pt}
\end{figure*} 

%% file: sections/method.tex
\vspace{-4pt}
\section{Method}
\label{method}
\vspace{-3pt}


Assuming the triplets~$(\mathcal{A, T, S})$ represent the audio, text, and sign space respectively, our goal is to learn the mapping from text or audio to sign within a unified framework~(Fig.~\ref{fig:framework}).
Given a training dataset~$\mathcal{D}\!=\!\{(\bm{a},\bm{t},\bm{s})\in\mathcal{A\times\! T\times\!S}\}$, MS2SL can realize text-to-sign $\mathcal{T\! \mapsto\! S}\!:\!\bm{s}\!=\!G(\bm{t})$ and audio-to-sign $\mathcal{A\! \mapsto\! S}\!:\!\bm{s}\!=\!G(\bm{a})$, where $G$ is the sign sequence diffusion generator.$_{\!}$
We initially employ pretrained models CLIP~\cite{DBLP:conf/icml/RadfordKHRGASAM21} and HuBERT~\cite{DBLP:journals/taslp/HsuBTLSM21} to$_{\!}$ extract$_{\!}$ features$_{\!}$ from$_{\!}$ text~$\bm{t}$$_{\!}$ and$_{\!}$ audio~$\bm{a}$.$_{\!}$ 
Next, we employ three encoders $E_a, E_t, E_s$ to encode these features, acquiring their embeddings$_{\!}$ $\bm{e}_a$,$_{\!}$ $\bm{e}_t$,$_{\!}$ and$_{\!}$ $\bm{e}_s$.$_{\!}$ 
~Subsequently, drawing on the operating mechanism of diffusion models, we employ a diffusion step encoder~$E_{h}$ and a sign noise encoder~$E_n$ to$_{\!}$ encode$_{\!}$ step~$h$$_{\!}$ and$_{\!}$ noise~$\bm{n}$ to $\bm{e}_h$ and $\bm{e}_n$$_{\!}$, respectively.
Finally,$_{\!}$ we$_{\!}$ utilize the generator~$G$ to produce the sign sequences: $\hat{\bm{s}}_t=G(\bm{e}_t, \bm{e}_{h}, \bm{e}_n)$ and $\hat{\bm{s}}_a=G(\bm{e}_a, \bm{e}_{h}, \bm{e}_n)$.

The paucity of co-occurring triplet data renders the direct training of MS2SL a formidable task.  
To overcome this challenge, we develop a joint embedding space that facilitates the natural alignment of multimodal data.
Furthermore, we employ ECL strategy to exploit the reciprocity among modalities within the embedding space, effectively furnishing feedback signals to boost the training. 
\subsection{Sign Predictor} \label{sec:SP}
\noindent\textbf{Cross-linguistic$_{\!}$ Modeling.}$_{\!}$
MS2SL$_{\!}$ aims$_{\!}$ to$_{\!}$ solve$_{\!}$ the$_{\!}$ problem$_{\!}$ of$_{\!}$ generating$_{\!}$ variable-length$_{\!}$ sequences$_{\!}$ across$_{\!}$ modalities.$_{\!}$
It$_{\!}$ necessitates phonological modeling between spoken and sign language, associating text and audio to the same target sign sequence.$_{\!}$
The$_{\!}$ causal$_{\!}$ attention$_{\!}$ mechanism can serve as a potent remedy for this challenging issue.$_{\!}$
Taking text-to-sign as an example, we first concatenate the embeddings of text~$\bm{e}_t$, denoising step~$\bm{e}_{h}$ and noise~$\bm{e}_n$.$_{\!}$
Next,$_{\!}$ we$_{\!}$ apply the causal self-attention~\cite{radford2018improving} to model the relationship among them.$_{\!}$
The \textit{mask} in causal attention ensures that the model only processes past and present information, maintaining temporal and logical coherence in the output.
As such, the output is computed as: $\text{CausalAtt}[\bm{e}_t;\bm{e}_{h};\bm{e}_n]$.
During inference, we initiate from the text embedding and produce indices autoregressively, ceasing generation when the model predicts the sequences.
Likewise, the concatenated entity of the audio~$\bm{e}_a$, step~$\bm{e}_{h}$, and noise~$\bm{e}_n$ can also undergo the causal attention to capture the relationship between audio and sign.
In causal attention, we adopt the common practice of positional encoding, which can model keypoints and inter-frame context while capturing cross-modal relations.
Thus, to simplify the model structure, we does not explicitly design a temporal module.
Finally, we employ two fully connected layers to output the sign prediction~$\hat{\bm{s}}_h$ for step $h$.

\noindent\textbf{Sign$_{\!}$ Language$_{\!}$ Production.}$_{\!}$
We$_{\!}$ apply$_{\!}$ a$_{\!}$ diffusion model as the sign generator.
Similarly, taking text-to-sign as an example, the diffusion generator $G$ is responsible for the gradually producing a continuous sign sequence~$\hat{\bm{s}}$. 
Diffusion generator $G$~simulates data distribution through a gradual forward and reversible process~\cite{DBLP:conf/nips/HoJA20},
training by maximizing the evidence lower bound to approximate target distributions.$_{\!}$ 
Diffusion model aims$_{\!}$ to$_{\!}$ reconstruct the input from a latent variable.
The forward process gradually transforms the input into noise by adding Gaussian noise. The reverse process starts from random noise and progressively removes the noise to recover the original data.

Common training for diffusion models involves independent$_{\!}$ noise$_{\!}$ prediction$_{\!}$ at$_{\!}$ each$_{\!}$ forward$_{\!}$ step~$h$,$_{\!}$ potentially$_{\!}$ reducing$_{\!}$ sequence$_{\!}$ coherence$_{\!}$ and$_{\!}$ consistency.
Following~\cite{DBLP:conf/cvpr/Shalev-Arkushin23}, we adopt the holistic training method.
We apply a schedule function $\small \delta_{h}=1/\text{log}{(h+1)}$ ($\delta \in [0,1]$) and a step size $\alpha_{h} = \delta_{h} - \delta_{h+1}$.
The predicted signs~$\hat{\bm{s}}_{h}$ at step~$h$, as: 
\vspace{-8pt}
\begin{equation} \label{eq:6}
\hat{s}_{h} = \alpha_{h} p_{h} + (1 - \alpha_{h}) \hat{s}_{h-1},
\vspace{-7pt}
\end{equation}
where the predicted signs $p_{h}$ at step $h$ are given as $G(\bm{t})$.
This method utilizes the output from the previous iteration as the input for the subsequent step, gradually reducing the step size as the process continues.
Each step combines previous outcomes with current predictions, reducing reliance on the initial noise.$_{\!}$
We$_{\!}$ also$_{\!}$ enhance training robustness by introducing$_{\!}$ a$_{\!}$ random$_{\!}$ noise$_{\!}$ to$_{\!}$ $\hat{s}_{h}$$_{\!}$ at$_{\!}$ each$_{\!}$ step.
Finally,$_{\!}$ the$_{\!}$ predicted$_{\!}$ initial$_{\!}$ sign~$\hat{s}_0$$_{\!}$ is$_{\!}$ outputted.
The$_{\!}$ loss$_{\!}$ of$_{\!}$ the$_{\!}$ diffusion$_{\!}$ is$_{\!}$ defined as:
\vspace{-5pt}
\begin{equation} \label{eq:7}
\mathcal{L}_{d} = \alpha_{h} s_{0} + (1 - \alpha_{h}) s_{{h}+1}.
\vspace{-1pt}
\end{equation}
\subsection{Modality Binding} \label{sec:MB}
%
MS2SL$_{\!}$ operates$_{\!}$ in$_{\!}$ an$_{\!}$ aligned embedding space, typically dependent on audio, text, and sign data for tri-modal alignment.
However, the difficulty for people with hearing impairments to perceive sound variations poses a challenge in recording these co-occurring triplets. 
Fortunately, ImageBind~\cite{DBLP:conf/cvpr/GirdharELSAJM23} reveals that a model can learn to align modalities in a joint embedding space by employing contrastive learning~\cite{DBLP:conf/cvpr/HadsellCL06}.
Training with (Image, Modality1) and (Image, Modality2) pairs can lead to a spontaneous alignment of Modality1 and Modality2 in embedding space. 
This alignment allows the model to excel in various tasks without requiring direct training on specific pairs of (Modality1, Modality2).

We extend the findings of ImageBind and construct a joint embedding space for the triplet dataset~$(\mathcal{A, T, S})$,
where MS2SL employs (text, sign) pairs as anchors to establish a cohesive space linking audio, text, and sign.
Let's explore a pair of modalities $(\mathcal{T}, \mathcal{S})$ with aligned observations.
Given a sign sequence $\bm{s}$ and its corresponding caption~$\bm{t}$.  
We first employ pretrained models CLIP~\cite{DBLP:conf/icml/RadfordKHRGASAM21} to extract textual features and encode them into normalized embeddings: $\bm{e}_t$ and $\bm{e}_s$.
Then, we leverages the paired modalities $(\mathcal{T}, \mathcal{S})$ to align the text with sign.
The corresponding encoders are optimized by InfoNCE~\cite{oord2018representation} loss~$\mathcal{L}_{\mathcal{T},~\mathcal{S}}$:
\vspace{-3pt}
\begin{equation} \small \label{eq:1}
    \mathcal{L}_{\mathcal{T},~\mathcal{S}} = -\text{log} \frac{\exp(\text{sim}(\bm{e}_t, \bm{e}_s) / \tau)}{\sum_{m=1}^{M} \exp(\text{sim}(\bm{e}_t, \bm{e}_{s_m}) / \tau)}.
\vspace{-3pt}
\end{equation}
Within$_{\!}$ the$_{\!}$ mini-batch,$_{\!}$ we$_{\!}$ consider$_{\!}$ each$_{\!}$ instance,$_{\!}$ whose$_{\!}$ index$_{\!}$ is$_{\!}$ not$_{\!}$ equal$_{\!}$ to$_{\!}$ $m$,$_{\!}$ as$_{\!}$ a$_{\!}$ negative$_{\!}$ example.
This$_{\!}$ approach$_{\!}$ aims$_{\!}$ to$_{\!}$ draw$_{\!}$ different$_{\!}$ embedding$_{\!}$ pairs$_{\!}$ closer$_{\!}$ within$_{\!}$ their$_{\!}$ joint$_{\!}$ embedding$_{\!}$ space.
Similarly,$_{\!}$ we$_{\!}$ can$_{\!}$ also$_{\!}$ obtain~$\mathcal{L}_{\mathcal{A,~S}}$~and~$\mathcal{L}_{\mathcal{T,~A}}$ for the pairs~$(\mathcal{A, S})$~and~$(\mathcal{T, A})$. 
Interestingly, we also observe the emergent alignment between modal pairs $(\mathcal{T}, \mathcal{A})$ in our embedding space. This phenomenon can occur when the training is solely based on pairs $(\mathcal{T}, \mathcal{S})$ and $(\mathcal{A}, \mathcal{S})$,  a trend that mirrors the findings reported in~\cite{DBLP:conf/cvpr/GirdharELSAJM23}.
Accordingly, MS2SL is designed to mainly leverage modal pairs $(\mathcal{T}, \mathcal{S})$ and $(\mathcal{T}, \mathcal{A})$, circumventing the need for triplet data.
In practice, this is achieved by employing a triadic loss:
\vspace{-5pt}
\begin{equation} \small \label{eq:1}
    \mathcal{L}_{nce} = \mathcal{L}_{\mathcal{T},~\mathcal{S}} + \mathcal{L}_{\mathcal{T},~\mathcal{A}} + \mathcal{L}_{\mathcal{A},~\mathcal{S}}.
\vspace{-5pt}
\end{equation}
As such, the embedding space can not only spontaneously align unseen triples but also be used in reconstructing unobserved modalities in ECL.
\vspace{-3pt}
\subsection{Embedding-consistency Learning} \label{sec:ECL}
Given a tuple $(\mathcal{A},\mathcal{T}, \mathcal{S})$, we employ a cyclic approach with the bound joint embedding to generate feedback signals for bidirectional cross-learning, fostering model training.
When triplet data is available, the encoders first extract features from their respective modalities.
Then, audio and text independently generate predicted sign language sequences~$\hat{\bm{s}}_a$~and~$\hat{\bm{s}}_t$.
To fully utilize real data, we calculate ECL loss after 500 epochs of model training.
The vanilla model, built on authentic data, guarantees minimal distribution differences between generated pseudo-embeddings and the original dataset.
Semantic consistency is calculated using the embeddings~$\hat{\bm{e}}_t$~and~$\hat{\bm{e}}_a$ from encoder~$E_s$, which encodes the two predicted sequences.
We can obtain the text-to-sign error~$\Delta(\hat{\bm{e}}_t,~\bm{e}_s)$ and the audio-to-sign loss~$\Delta(\hat{\bm{e}}_a,~\bm{e}_s)$:
\vspace{-5pt}
\begin{equation} \normalsize \label{eq:8}
\begin{aligned}
 \Delta(\hat{\bm{e}}_t,~\bm{e}_s)\!=\!\|\hat{\bm{e}}_t,\bm{e}_s\|_2,\\
 \Delta(\hat{\bm{e}}_a,~\bm{e}_s)\!=\!\|\hat{\bm{e}}_a,\bm{e}_s\|_2.
\end{aligned}
\vspace{-6pt}
\end{equation}
Evaluation scores are derived from comparing the two embeddings~$\hat{\bm{e}}_t$~and~$\hat{\bm{e}}_a$.
Both audio and text can receive feedback signals from the generative streams of each other.
To compensate for the missing audio modality and ensure smooth processing, we use a mapping network $M$ and text embeddings to generate pseudo audio features. 
The operation is conducted in the embedding space, thus minimally affecting inference speed.
For unpaired natural audios~$\mathcal{U}$, we can get the formula:
\vspace{-4pt}
\begin{equation} \small \label{eq:8}
\begin{aligned}
\!\!\!\mathcal{L}_{(\mathcal{T},~\mathcal{A},~\mathcal{S})} &\!=\! \|E_s(G(\bm{e}_a))\!-\!E_s(G(\bm{e}_t))\|_2,\!\!\!\\
\vspace{8pt}
\!\!\!\mathcal{L}_{(\mathcal{T}',~\mathcal{S}')}~~ &\!=\! \|E_s(M(G(\bm{e}'_t)))\!-\!E_s(G(\bm{e}'_t))\|_2.\!\!\!
\end{aligned}
\vspace{-3pt}
\end{equation}
Then our ECL loss is defined as:
\vspace{-8pt}
\begin{equation} \label{eq:9}
\mathcal{L}_{ecl}\!=\!\mathcal{L}_{(\mathcal{T},~\mathcal{A},~\mathcal{S})~\!\in~\!\mathcal{D}}\! + \mathcal{L}_{(\mathcal{T}',~\mathcal{S}')~\!\in~\!\mathcal{U}}.
\vspace{-6pt}
\end{equation}
MS2SL translates entire spoken sentences into continuous sign language sequences.
Overall, our total loss comprises three components, \ie, the diffusion model loss, ECL loss, and joint embedding loss:
\vspace{-6pt}
\begin{equation} \label{eq:10}
\begin{aligned}
\mathcal{L} = \lambda_1\mathcal{L}_d + \lambda_2\mathcal{L}_{ecl}  + \lambda_3\mathcal{L}_{nce}, 
\end{aligned}
\vspace{-5pt}
\end{equation}
where the cofficients are empirically set as $\lambda_1=\lambda_2=\lambda_3=1$.
\vspace{-3pt}
\subsection{Implementation Details}
\vspace{-1pt}
\noindent\textbf{Training.}$_{\!}$
MS2SL$_{\!}$ takes$_{\!}$ speech audio$_{\!}$ or text as inputs.
We utilize pre-trained models for encoding both speech and text, HuBert~\cite{DBLP:journals/taslp/HsuBTLSM21} for speech and CLIP~\cite{DBLP:conf/icml/RadfordKHRGASAM21} for text.
We first extract embeddings~$\bm{e}_t, \bm{e}_a, \bm{e}_s, \bm{e}_h, \bm{e}_n$ through five encoders.
We employ keypoints to represent signs, like the 137 human keypoints in How2Sign~\cite{DBLP:conf/cvpr/DuartePVGDMTG21}, which are normalized and standardized before being input into the model.
$\bm{e}_t$, $\bm{e}_a$ and $\bm{e}_s$ participate in learning the joint embedding space.
Concurrently, $\bm{e}_t, \bm{e}_a, \bm{e}_h$ and $\bm{e}_n$ serve as conditions to control the generation of text-to-sign and audio-to-sign, respectively.
Here, we adopt the common practice~\cite{DBLP:journals/ijcv/SaundersCB21,SaundersCB21,DBLP:conf/cvpr/Shalev-Arkushin23} of using the first sign pose as initial noise.
The first 500 epochs skip the audio-to-sign generation flow in the absence of audio.
After obtaining a vanilla model, we apply the mapping network~$M$ to transform $\bm{e}_t$~into~$\bm{e}_a$ to continue the training until the model converges.
Since PHOENIX~\cite{DBLP:conf/lrec/ForsterSKBN14} dataset is in \textit{German}~sign language, and our pre-trained model is primarily based on \textit{English}, we utilize the penultimate layer features of CLIP along with MLP to align and transform between German and English.
As for ECL, we incorporate cycles among the three modalities, namely audio-to-sign, text-to-sign, and audio-to-text, greatly enhancing the efficiency of data utilization.
We adopt the commonly used exponential moving average~\cite{DBLP:conf/cvpr/CaiRMFTS21} strategy with diffusion parameters~\cite{DBLP:conf/cvpr/CaiRMFTS21} to ensure smoother, more robust training.
For details, please refer to the supplementary.

\noindent\textbf{Inference.}
The model can perform SLP from audio or text independently.
Inference for each modality involves executing the sequence sampling of the diffusion model.
Taking text-to-sign as an example, the process starts with CLIP encoding the text into features.
These text features are then fed into the sign predictor, which sequentially generates a sequence noise prediction.
The completion of this sampling process results in the generation of the desired sign sequence.
The process for generating signs from speech is similar.
We take the average of twenty generations to mitigate deviation.

\noindent\textbf{Reproducibility.}$_{\!}$
Our$_{\!}$ method$_{\!}$ is$_{\!}$ implemented$_{\!}$ using$_{\!}$ PyTorch$_{\!}$ on$_{\!}$ $2$$_{\!}$ RTX$_{\!}$ 4090$_{\!}$ GPUs,$_{\!}$ with$_{\!}$ a$_{\!}$ training$_{\!}$ time$_{\!}$ of$_{\!}$ about$_{\!}$ $12$$_{\!}$ hours$_{\!}$ and$_{\!}$ an$_{\!}$ average$_{\!}$ inference$_{\!}$ time$_{\!}$ of$_{\!}$ $0.3$$_{\!}$ seconds.$_{\!}$
Following$_{\!}$~\cite{DBLP:conf/cvpr/ZhangZCZZLSY23},$_{\!}$ we$_{\!}$ remove$_{\!}$ data$_{\!}$ with$_{\!}$ word$_{\!}$ count$_{\!}$ exceeding$_{\!}$ $20$. 

%% file: sections/experiment.tex
\begin{table*}[th]
    \centering
        \vspace{-8pt}
    \setlength\tabcolsep{1pt}
	\renewcommand\arraystretch{1.2}
    \resizebox{1\linewidth}{!}{

    \begin{tabular}{l c c c c c p{0mm} c c c c c c c}
    \thickhline
    \multirow{2}{*}{Methods}  & \multicolumn{5}{c}{How2Sign} & \multicolumn{1}{c}{} & \multicolumn{5}{c}{PHOENIX14T} \\
    \cline{2-6} \cline{8-12}
    ~ & BLEU-4 & BLEU-3 & BLEU-2 & BLEU-1 & ROUGE & & BLEU-4 & BLEU-3 & BLEU-2 & BLEU-1 & ROUGE \\


        Back-translation & \et{10.89}{.003} & \et{13.32}{.01} & \et{16.71}{.06} & \et{22.38}{.05} & \et{23.23}{.03} & & \et{20.53}{.01} & \et{25.13}{.03} & \et{32.81}{.04} & \et{44.01}{.02} & \et{45.61}{.03} \\
    \hline
        PT~\cite{DBLP:conf/eccv/SaundersCB20} & \et{2.01}{.02} & \et{3.86}{.04} & \et{7.04}{.00} & \et{13.69}{.04} & \et{13.81}{.03} & & \et{11.32}{.02} & \et{12.91}{.01} & \et{19.04}{.05} & \et{31.36}{.01} & \et{32.46}{.01}  \\
        MOMP~\cite{SaundersCB21} & \et{2.34}{.04} & \et{3.92}{.01} & \et{7.63}{.02} & \et{13.68}{.06} & \et{13.83}{.05} & & \et{11.19}{.03} & \et{13.14}{.02} & \et{19.64}{.01} & \et{32.22}{.04} & \et{32.96}{.02} \\
        Ham2Pose~\cite{DBLP:conf/cvpr/Shalev-Arkushin23} & \et{2.93}{.06} & \et{4.07}{.04} & \et{7.31}{.02} & \et{12.38}{.03} & \et{13.29}{.01} & & \et{11.71}{.03} & \et{13.22}{.03} & \et{20.16}{.05} & \et{33.39}{.00} & \et{34.02}{.04}  \\


        T2M-GPT~\cite{DBLP:conf/cvpr/ZhangZCZZLSY23} & \et{3.53}{.03} & \et{5.14}{.01} & \et{7.92}{.05} & \et{12.87}{.05} & \et{13.99}{.03} & & \et{11.66}{.02} & \et{13.35}{.07} & \et{21.19}{.00} & \et{35.24}{.02} & \et{35.44}{.03}  \\

    \hline
         MS2SL \textit{w/o} ECL & \etbb{3.76}{.02} & \etbb{6.03}{.02} & \etbb{8.05}{.04} & \etbb{14.51}{.05} & \etbb{15.10}{.06} & & \etbb{12.03}{.02} & \etbb{14.32}{.04} & \etbb{21.72}{.03} & \etbb{35.36}{.06} & \etbb{35.68}{.08} \\
        MS2SL-T2S  & \etr{\mathbf{4.26}}{.04} & \etr{\mathbf{6.84}}{.02} & \etr{\mathbf{9.17}}{.05} & \etr{\mathbf{14.67}}{.03} & \etr{\mathbf{16.38}}{.03} & & \etr{\mathbf{12.77}}{.06} & \etr{\mathbf{15.81}}{.07} & \etr{\mathbf{22.04}}{.03} & \etr{\mathbf{36.41}}{.01} & \etr{\mathbf{36.63}}{.03} \\
    \hline
    \end{tabular}
    \vspace{-8pt}
    }
    \vspace{-2mm}
    \caption{\textbf{Comparisons of text-to-sign with the state-of-the-art methods~(\S\ref{sec:CS}) on How2Sign~\cite{DBLP:conf/cvpr/DuartePVGDMTG21} and PHOENIX14T~\cite{DBLP:conf/cvpr/CamgozHKNB18}.} For each metric, we repeat the evaluation 20 times and report the average. \textcolor{red}{Red} and \textcolor{blue}{Blue} indicate the best and the second best result, respectively.}
    \label{tbl:CS}
    \vspace{-12pt}
\end{table*}
\section{Experiments}
\vspace{-3pt}
We evaluate the effectiveness of MS2SL under text-to-sign and audio-to-sign settings.
\subsection{Experimental Setup}
\label{sec:ES}
\vspace{-3pt}
\noindent\textbf{Datasets.} We conduct experiments on two continuous sign language datasets: 
\vspace{-7pt}
\begin{itemize} [leftmargin=*]
\item How2Sign~\cite{DBLP:conf/cvpr/DuartePVGDMTG21} is a challenging multimodal American sign language dataset with a 16k-word vocabulary and comprehensive annotations.
    It includes $1,176$ entries with audio and has train/dev/test splits of $31165/1741/2357$.
\vspace{-10pt}
\item PHOENIX14T~\cite{DBLP:conf/cvpr/CamgozHKNB18}, a widely applied German weather sign language dataset, contains $2,887$ words, $1,066$ sign annotations, with train/dev/test splits of $7096/519/642$.
\end{itemize}
\vspace{-7pt}
\noindent\textbf{Evaluation Metrics.}
Following~\cite{DBLP:conf/eccv/SaundersCB20}, we adopt back-translation approach for evaluating, \ie, we leverage the cutting-edge SLT model~\cite{DBLP:conf/cvpr/CamgozKHB20} to ingeniously translate back from generated signs to text.
Subsequently, we calculate BLEU~\cite{DBLP:conf/acl/PapineniRWZ02} and ROUGE~\cite{lin2004rouge} scores, which are commonly used metrics for SLP and machine translation.
We apply ROUGE-L F1-Score and report BLEU-1 to BLEU-4 for translation performance at different phrase lengths.

\noindent\textbf{Competitors.} For text-to-sign generation stream, we consider four SOTA competitors:
\vspace{-6pt}
\begin{itemize} [leftmargin=*]
  \item Ham2Pose~\cite{DBLP:conf/cvpr/Shalev-Arkushin23}, which employs transformer and diffusion model, animates HamNoSys (a sign notation) into sign poses. 
  \vspace{-7pt}
  \item T2M-GPT~\cite{DBLP:conf/cvpr/ZhangZCZZLSY23} combines VQ-VAE~\cite{DBLP:conf/nips/OordVK17} and CLIP~\cite{DBLP:conf/icml/RadfordKHRGASAM21} for motion generation.
  \vspace{-7pt}
  \item PT~\cite{DBLP:conf/eccv/SaundersCB20} translates discrete spoken sentences into sign sequences. 
  \vspace{-7pt}
  \item MOMP~\cite{SaundersCB21} divides SLP into two sub-tasks: latent sign representation and animation imitation. 
\end{itemize}
\vspace{-6pt}
\noindent{As for the audio-to-sign stream, since there are not specific methods, we extend MS2SL to multiple implementations for a thorough evaluation, including audio-to-sign, audio-to-text-to-sign, and text-to-audio-to-sign. 
For audio-to-text-sign, we apply WeNet~\cite{YaoWWZYYPCXL21} to translate audio into text, followed by the generation of signs. Conversely, for text-to-audio-to-sign, we employ DeepVoice~\cite{GibianskyADMPPR17} to convert text into audio for subsequent sign generation.}
\vspace{-4pt}
\subsection{Comparison to State-of-the-art} \label{sec:CS}
\vspace{-3pt}
\noindent\textbf{Quantitative$_{\!}$ Results.} We present$_{\!}$ the$_{\!}$ comparative$_{\!}$ analysis$_{\!}$ results$_{\!}$ in Table~\ref{tbl:CS} on How2Sign and PHOENIX14T test set.
MS2SL demonstrates impressive gains against the four robust methods, establishing a new benchmark for SOTA performance.
In the generation of text-to-sign, our approach yields a ROUGE of $14.67$, marking a notable increase of $2.39$ over its counterpart (T2M-GPT, which has a $13.99$ ROUGE).
Furthermore, MS2SL combined with ECL surpasses the standalone by $1.28$.
How2Sign~\cite{DBLP:conf/cvpr/DuartePVGDMTG21} and PHOENIX14T~\cite{DBLP:conf/cvpr/CamgozHKNB18} are datasets of different scales, demonstrating the robustness of our method and the burgeoning potential of diffusion models in generating long sign sequences.


Table~\ref{tbl:A2S} reports the audio-to-sign results on How2Sign, noting that PHOENIX14T is not included here due to the absence of audio data.
Our method significantly enhance performance, achieving notable improvements (\ie, BLEU-1 increase from $9.49$ to $11.77$, ROUGE from $9.60$ to $12.16$).
The ECL strategy also enhances ROUGE by $1.12$.
Considering the scarcity of audio modality data, this achievement is particularly noteworthy and shows its real-world applicability.
We can also conclude that it is difficult to obtain a well-performing model by training solely with the limited audio data in How2Sign. 
This also highlights the urgency of utilizing non-co-occurring triplets. 
\begin{table}[t]
\vspace{3pt}
\centering
    \resizebox{1.\linewidth}{!}{
                \setlength\tabcolsep{1pt}
                \renewcommand\arraystretch{1.28}
    \begin{tabular}{c||ccccc}
        \thickhline
           Methods & BLEU-4 & BLEU-3 & BLEU-2 & BLEU-1 & ROUGE \\
           \hline
            T2A2S~(\S\ref{sec:ES}) &  \et{0.98}{.04} & \et{1.32}{.02} & \et{3.71}{.02} & \et{8.38}{.01} & \et{8.52}{.00} \\
            A2T2S~(\S\ref{sec:ES}) &  \et{1.02}{.02} & \et{1.47}{.01} & \et{4.66}{.05} & \et{9.49}{.08} & \et{9.60}{.04} \\
            \cdashline{1-6}
            MS2SL \textit{w/o} ECL &  \et{1.24}{.07} & \et{1.63}{.03} & \et{4.71}{.01} & \et{10.59}{.01} & \et{11.04}{.03} \\
            MS2SL-A2S &  \etr{\mathbf{1.67}}{.01} & \etr{\mathbf{1.94}}{.03} & \etr{\mathbf{5.90}}{.02} & \etr{\mathbf{11.77}}{.05} & \etr{\mathbf{12.16}}{.01} \\
            \hline
    \end{tabular}
    }
    \vspace{-8pt}
    \caption{\textbf{Audio-to-Sign results on How2Sign~(\S\ref{sec:CS}).}}
    \label{tbl:A2S}
    \vspace{-17pt}
\end{table}
\input{figs/results_example}

\noindent\textbf{Qualitative$_{\!}$ Comparison.}$_{\!}$
Fig.~\ref{fig:results_example}$_{\!}$ presents visual results on How2Sign~\cite{DBLP:conf/cvpr/DuartePVGDMTG21}.
It demonstrates that our method can produce signs that are more closely aligned with their semantic meaning.
After meticulous examination, it is evident that MS2SL surpasses other models in generating actions with smoother transitions, heightens expressiveness, greater diversity, and superior adherence to physical constraints.
Some noise and jitter are noted in the audio-to-sign generation stream. 
The main reason is that our method focuses on translating complete spoken content into sign sequences, whereas previous studies~\cite{DBLP:conf/eccv/SaundersCB20,DBLP:journals/ijcv/SaundersCB21,DBLP:conf/cvpr/Shalev-Arkushin23} target the creation of discrete lexical symbol or phrase. 
The challenge of training models to convey extended semantic content and long sequences often leads to incoherent movements during sign generation.  

\begin{table}[t]
\vspace{3pt}
\centering
    \resizebox{0.99\linewidth}{!}{
                \setlength\tabcolsep{5pt}
                \renewcommand\arraystretch{1.05}
    \begin{tabular}{l||cc}
        \thickhline
           Methods & How2Sign & PHOENIX14T \\
           \hline
          PT~\cite{DBLP:conf/eccv/SaundersCB20}  & 1.29  &  1.54 \\
          Ham2Pose~\cite{DBLP:conf/cvpr/Shalev-Arkushin23} & 1.97 &  1.73  \\
          A2T2S~(\S\ref{sec:ES})  & 1.87  &  2.09 \\
          T2M-GPT~\cite{DBLP:conf/cvpr/ZhangZCZZLSY23} & 2.19 &  2.20   \\
              \hline
          MS2SL  &  $\textcolor{red}{\mathbf{2.65}}$ & $\textcolor{red}{\mathbf{3.21}}$ \\
            \hline
    \end{tabular}
    }
    \vspace{-8pt}
    \caption{\textbf{User study~(\S\ref{sec:CS}).}}
    \label{tab:US}
    \vspace{-17pt}
\end{table}
\noindent\textbf{User Study.}
Given the challenge of finding sign language experts, who require extensive training, we conduct a user study with $10$ hearing volunteers.
We ask the volunteers to compare sign sequences generated by  different methods. 
We slow down sign sequence playback for easier comparison by volunteers.
Volunteers select the sequence closer to the ground truth and assign a score. Our scoring range is from $1$ to $5$, with higher scores indicating closer proximity to the ground truth.
Most participants report that the sign sequences generated by MS2SL are smoother and more accurate~(Table~\ref{tab:US}).
User feedback highlight the advantages of MS2SL in terms of expression clarity and pose accuracy.
\begin{table*}[h]
    \vspace*{-1pt}
    \begin{minipage}{\textwidth}
        \begin{subtable}[t]{0.48\textwidth}
            \makeatletter\def\@captype{table}\captionsetup{width=1.\linewidth}
             \vspace{-3pt}
            \centering\small
            \resizebox{1.\linewidth}{!}{
                \setlength\tabcolsep{4pt}
                \renewcommand\arraystretch{1.43}
                \begin{tabular}{c||ccccc}
                \thickhline
                   Methods & BLEU-4 & BLEU-3 & BLEU-2 & BLEU-1 & ROUGE \\
                   \hline
                   Audio &  \et{0.98}{.04} & \et{1.32}{.02} & \et{3.71}{.02} & \et{8.38}{.01} & \et{8.52}{.00} \\
                   Text &  \et{1.74}{.00} & \et{2.41}{.02} & \et{3.43}{.07} & \et{8.62}{.03} & \et{9.57}{.01} \\
                   T2A2S &  \et{1.85}{.0} & \et{2.35}{.03} & \et{4.26}{.02} & \et{8.52}{.08} & \et{9.28}{.03} \\
                    \cdashline{1-6}
                   MS2SL &  \etr{4.26}{.04} & \etr{6.84}{.02} & \etr{9.17}{.05} & \etr{14.67}{.03} & \etr{16.38}{.03} \\
                    \hline
                \end{tabular}
            }
            \vspace{-1pt}
            \caption{data from different modalities} 
            \label{tbl:DDM}
        \end{subtable}
        \hspace{2ex}
        \begin{subtable}[t]{0.48\textwidth}
            \makeatletter\def\@captype{table}\captionsetup{width=.9\linewidth}
            \vspace{-3pt}
            \centering\small
            \resizebox{1.\linewidth}{!}{
                \setlength\tabcolsep{3pt}
                \renewcommand\arraystretch{1.35}
                \begin{tabular}{c||ccccc}
                \thickhline
                   Methods & BLEU-4 & BLEU-3 & BLEU-2 & BLEU-1 & ROUGE \\
                   \hline
                    $0k$ &  \et{3.76}{.06} & \et{6.03}{.02} & \et{8.05}{.05} & \et{14.51}{.04} & \et{15.10}{.02} \\
                    $5k$ &  \et{3.79}{.06} & \et{6.23}{.02} & \et{8.17}{.05} & \et{14.62}{.04} & \et{15.56}{.02} \\
                    $10k$ &  \et{3.82}{.03} & \et{6.37}{.03} & \et{8.31}{.02} & \et{14.57}{.06} & \et{15.87}{.00} \\
                    $15k$ &  \etr{4.26}{.04} & \etr{6.84}{.02} & \etr{9.17}{.05} & \etr{14.67}{.03} & \etr{16.38}{.03} \\
                    \hline
            \end{tabular}
            }
            \vspace{-1pt}
            \caption{embedding consistency learning} 
            \label{tbl:ECL}
        \end{subtable}
    \end{minipage}

    \vspace{7pt}

    \begin{minipage}{\textwidth}
        \begin{subtable}[t]{0.48\textwidth}
            \makeatletter\def\@captype{table}\captionsetup{width=1.\linewidth}
            \centering\small
            \resizebox{1.\linewidth}{!}{
                \setlength\tabcolsep{5pt}
                \renewcommand\arraystretch{1.2}
                \begin{tabular}{c||ccccc}
                \thickhline
                   Steps & BLEU-4 & BLEU-3 & BLEU-2 & BLEU-1 & ROUGE \\
                   \hline
                   0 &  \et{0.62}{.02} & \et{2.08}{.03} & \et{4.16}{.07} & \et{9.57}{.03} & \et{9.72}{.05} \\
                   5 &  \et{1.09}{.04} & \et{2.42}{.06} & \et{5.24}{.04} & \et{10.44}{.00} & \et{10.90}{.01} \\
                   10&  \etbb{4.26}{.04} & \etr{6.84}{.02} & \et{9.17}{.05} & \et{14.67}{.03} & \et{16.38}{.03} \\
                   15 &  \et{4.04}{.01} & \et{6.23}{.04} & \etbb{9.58}{.0} & \etbb{15.26}{.02} & \etr{17.33}{.01} \\
                   20 &  \etr{4.87}{.01} & \etbb{6.66}{.04} & \etr{9.67}{.0} & \etr{15.45}{.02} & \etbb{17.24}{.01} \\
                    \hline
                \end{tabular}

            }
            \vspace{-1pt}
            \caption{diffusion model} 
            \label{tbl:DM}
        \end{subtable}
        \hspace{2ex}
        \begin{subtable}[t]{0.48\textwidth}
            \makeatletter\def\@captype{table}\captionsetup{width=1.\linewidth}
            \centering\small
            \resizebox{1.\linewidth}{!}{
                \setlength\tabcolsep{1pt}
                \renewcommand\arraystretch{1.63}
                \begin{tabular}{l||ccccc}
                \thickhline
                   Pre-trained & BLEU-4 & BLEU-3 & BLEU-2 & BLEU-1 & ROUGE \\
                   \hline
                   WavLM~\cite{DBLP:journals/jstsp/ChenWCWLCLKYXWZ22} &  \et{1.63}{.06} & \et{1.79}{.02} & \et{6.12}{.01} & \et{10.94}{.00} & \et{11.43}{.02} \\
                   HuBert~\cite{DBLP:journals/taslp/HsuBTLSM21} &  \et{1.67}{.07} & \et{1.94}{.04} & \et{5.90}{.02} & \et{11.77}{.06} & \et{12.16}{.01} \\
                   \cdashline{1-6}
                   CLIP~\cite{DBLP:conf/icml/RadfordKHRGASAM21} &  \etr{4.26}{.04} & \et{6.84}{.02} & \et{9.17}{.05} & \etr{14.67}{.03} & \et{16.38}{.03} \\
                   BERT~\cite{DBLP:conf/naacl/DevlinCLT19} &  \et{4.11}{.04} & \etr{6.91}{.02} & \etr{10.27}{.01} & \et{13.37}{.05} & \etr{16.52}{.06} \\
                    \hline
            \end{tabular}
            }
            \vspace{-1pt}
            \caption{different pretrained models} 
            \label{tbl:DPM}

        \end{subtable}
    \end{minipage}
    \vspace{-4pt}
    \caption{\textbf{A set of ablation studies~(\S\ref{exp:AS}).} All experiments employ the same network and structure, with slight variations arising due to different inputs. We report the results of text-to-sign generation by default.}
    \vspace*{-10pt}
\end{table*}
\vspace{-3pt}
\subsection{Ablation Study} \label{exp:AS}
\vspace{-3pt}
We conduct careful profiling of the impact of each module within MS2SL on How2Sign.

\noindent\textbf{Data in Different Modalities.}
We primarily conduct four experiments: audio-to-sign, text-to-sign, text-to-audio-to-sign, and MS2SL, to compare and analyze the role of different modalities.
As shown in Table~\ref{tbl:DDM}, although direct generation from audio-to-sign and text-to-sign can yield appropriate results, MS2SL significantly outperforms them.
Removal of text data leads to a $6.29$ decrease in BLEU-1, highlighting its crucial role.$_{\!}$
The mediating role of text leads to an increase~$0.76$ in ROUGE. 
Multimodal data yields superior results compared to its unimodal counterpart, enriching the learning process with more diverse information.

\noindent\textbf{Embedding Consistency Learning.}
We investigate the impact of the cyclical consistency training presented in~\S~\ref{sec:ECL}, and the results are illustrated in Table~\ref{tbl:ECL}.
We note that common training method performs comparably to baseline models, while cyclical consistency boosts model performance akin to adding substantial training data.
Compared to the alternative only with single modality, MS2SL approach shows a $1.12$ increase in BLEU-2 and a $1.28$ increase in ROUGE, demonstrating the synergistic effect of integrating data from multiple modalities.
We further pay particular attention to the impact of dataset size.
We also observe a direct correlation between dataset size and model accuracy. 
For smaller datasets (under $10k$ samples), the accuracy plateau around $15.5$.
Several insights can be drawn:
i) Performances improve as more training data is used.
ii) Over $10k$ unpaired data entries, the signs might be of good quality, but the model cannot further improve on a large scale, possibly due to the scarcity of audio. 
This trend shows that more data notably improves sequence generation, even without clear semantic boundaries.

\noindent\textbf{Diffusion Model.}
As shown in~Table~\ref{tbl:DM}, implementing the diffusion model lead to a significant enhancement.
The quality metrics, such as BLEU-1 and ROUGE, improved by $5.1$ and $6.66$, respectively, compared to non-diffusion model approach.
Our study explores denoising steps ranging from $5$ to $20$, revealing a discernible trade-off between generation quality and computational efficiency.
Compared to a fixed 10-step denoising process, the 20-step process unsteadily improve $0.78$ in BLEU-1 by approximately $5.3\%$ with a disproportionate increase in computational load.
Thus, in this paper, $10$ is set as the default number of denoising steps.

\noindent\textbf{Pre-trained Models.}
We select four widely used models, including, CLIP~(text), BERT~(text)~\cite{DBLP:conf/naacl/DevlinCLT19}, HuBert~(audio) and WavLM~(audio)~\cite{DBLP:journals/jstsp/ChenWCWLCLKYXWZ22}, to assess their impact on performance.
As shown in Table~\ref{tbl:DPM}, for audio-to-sign generation, the impact of HuBert and WavLM on performance is minor, with negligible differences observed between the two pre-trained models.
GPT outperforms CLIP models in text-related tasks, with a slight improvement of up to $0.14$ in ROUGE. 
This may be because BERT focuses on natural language processing, leading to enhanced text understanding capabilities.
\vspace{-3pt}

%% file: figs/results_example.tex
\begin{figure*}[t]

    \centering

    \setlength{\tabcolsep}{0.1pt}

    \renewcommand{\arraystretch}{0.5}

    \resizebox{.9\textwidth}{!}{
    \begin{tabular}{c@{\hspace{0.3em}}c@{\hspace{0.3em}}c@{\hspace{0.3em}}c@{\hspace{0.3em}}c@{\hspace{0.6em}}c@{\hspace{0.3em}}c@{\hspace{0.3em}}c@{\hspace{0.3em}}c@{\hspace{0.3em}}c}
        
        \rule{0pt}{7mm}\rotatebox{90}{{\small ~~Text}} & \multicolumn{4}{c}{\includegraphics[width=0.35 \linewidth, trim=25bp 20bp 20bp 20bp, clip]{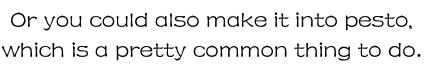}} & \rotatebox{90}{\small Audio}&
            \multicolumn{4}{c}{\includegraphics[width=59mm, height=7mm, trim=10bp 10bp 10bp 10bp, clip]{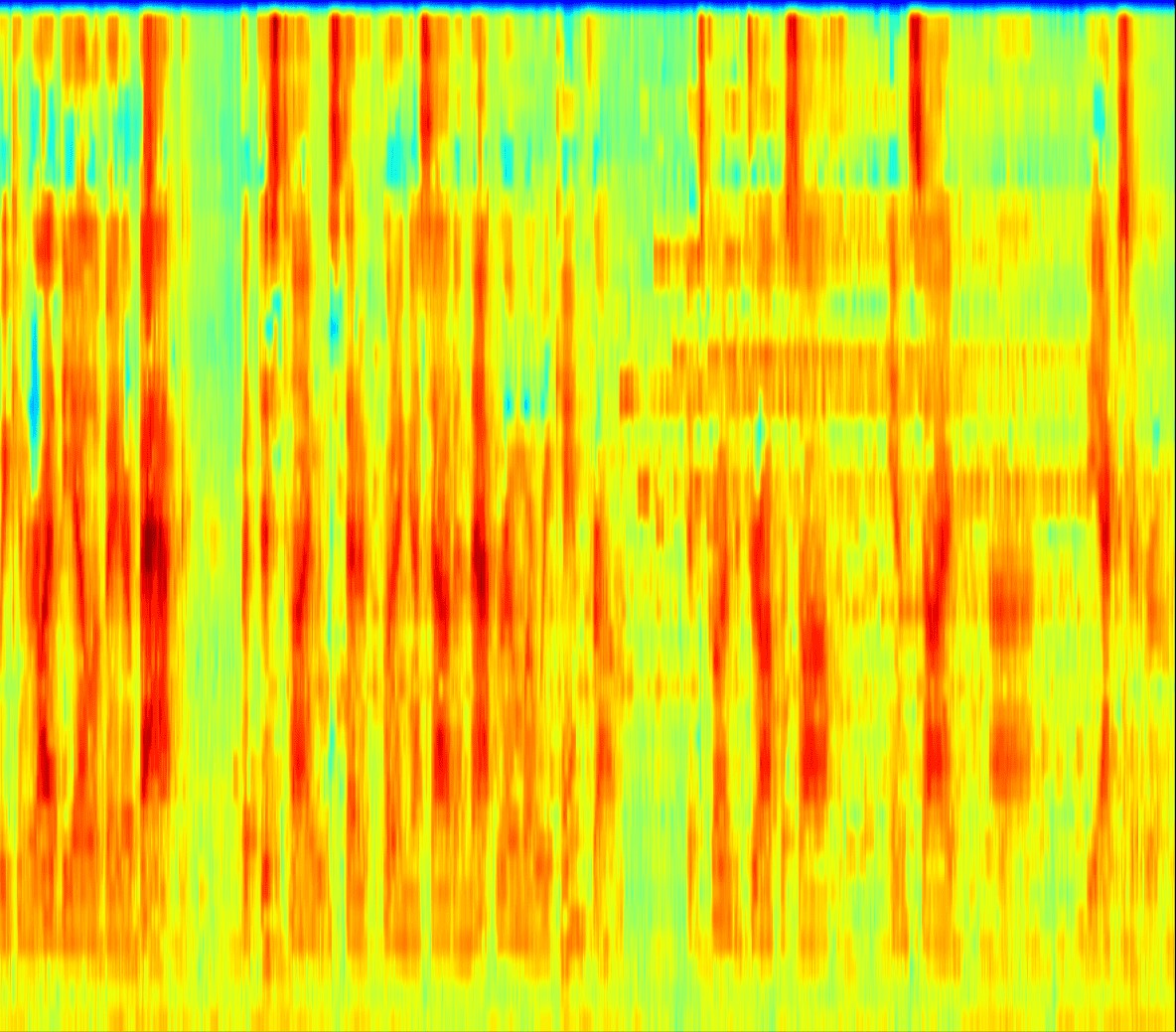}} \\ 
        
        \rotatebox{90}{\phantom{Aa.}{\small Video}} &

        \includegraphics[width=14mm, height=17mm]{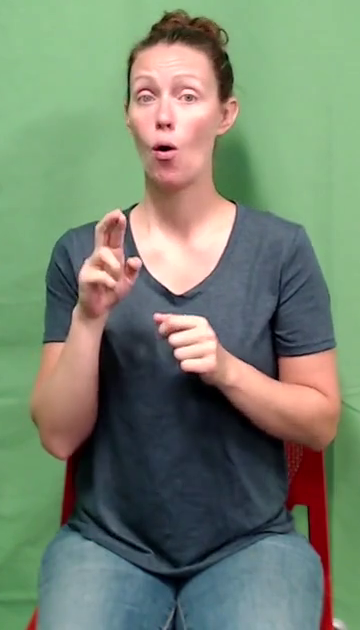} &

        \includegraphics[width=14mm, height=17mm]{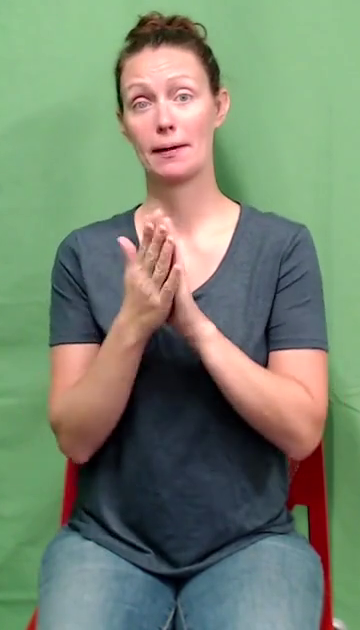} &

        \includegraphics[width=14mm, height=17mm]{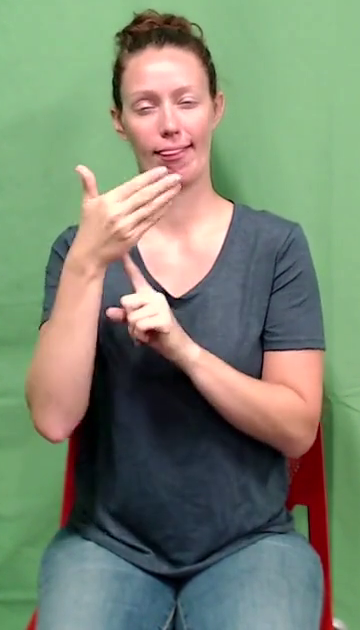} &

        \includegraphics[width=14mm, height=17mm]{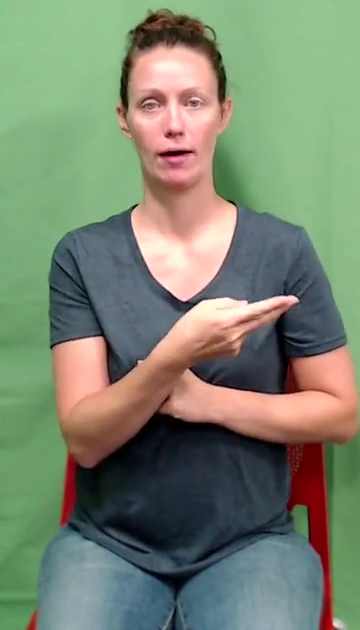} &
        
        \rotatebox{90}{\phantom{Aa.}{\small Video}} &
        
        \includegraphics[width=14mm, height=17mm]{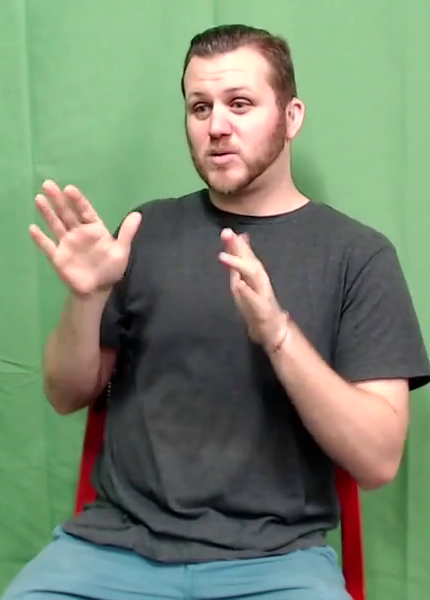} &

        \includegraphics[width=14mm, height=17mm]{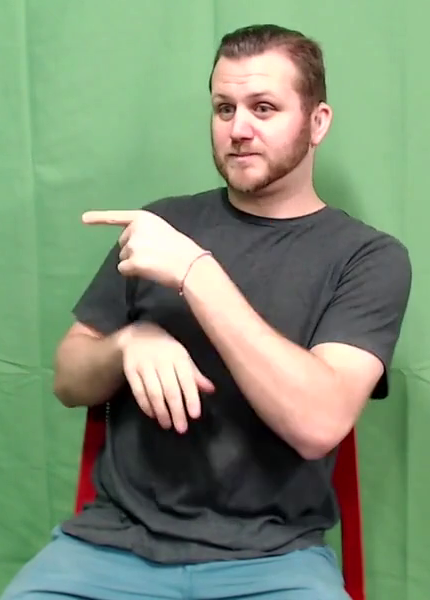} &

        \includegraphics[width=14mm, height=17mm]{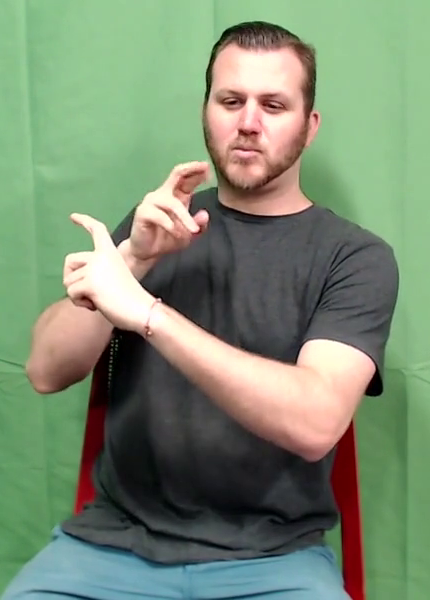} &
        
        \includegraphics[width=14mm, height=17mm]{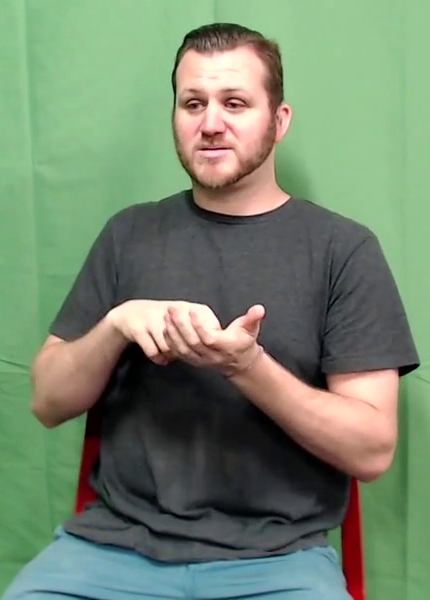}  \\

        \rotatebox{90}{\phantom{Aaa}{\small GT}} &

        \includegraphics[width=12.7mm, height=16mm]{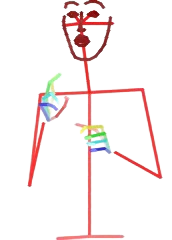} &

        \includegraphics[width=13.2mm, height=16mm]{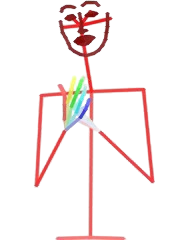} &

        \includegraphics[width=13.2mm, height=16mm]{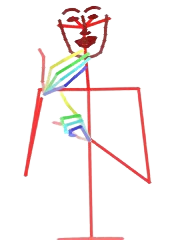} &

        \includegraphics[width=12.7mm, height=16mm]{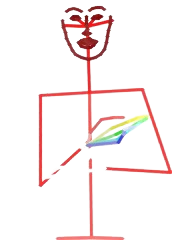} &
        
        \rotatebox{90}{\phantom{Aaa}{\small GT}}&
        
        \includegraphics[width=11.7mm, height=16mm]{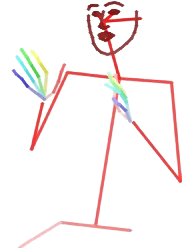} &

        \includegraphics[width=12.5mm, height=16mm]{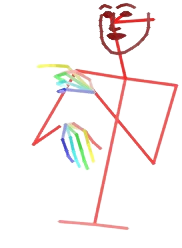} &

        \includegraphics[width=12.5mm, height=16mm]{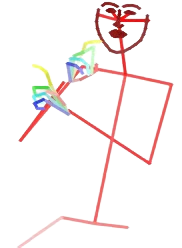} &
        
        \includegraphics[width=11.7mm, height=16mm]{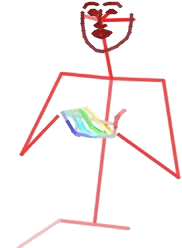}  \\

       \rotatebox{90}{\phantom{A}{\small \textbf{MS2SL}}} &

        \includegraphics[width=12mm, height=16mm]{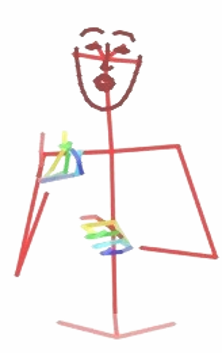} &

        \includegraphics[width=12mm, height=16mm]{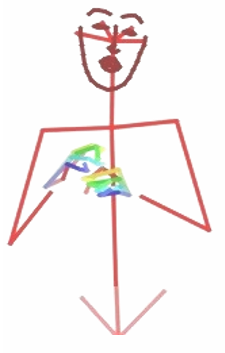} &

        \includegraphics[width=12.5mm, height=16mm]{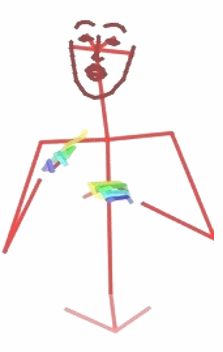} &

        \includegraphics[width=11mm, height=16mm, trim=1bp 1bp 3bp 1bp, clip]{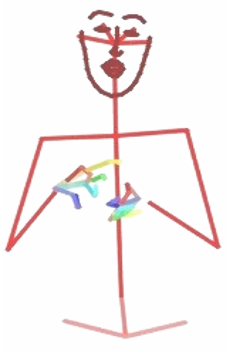} &
        
        \rotatebox{90}{\phantom{A}{\small \textbf{MS2SL}}} &
        
        \includegraphics[width=11mm, height=16mm]{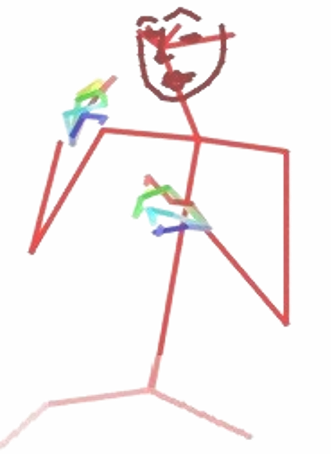} &

        \includegraphics[width=13mm, height=17.5mm, trim=1bp 22bp 0bp 1bp, clip]{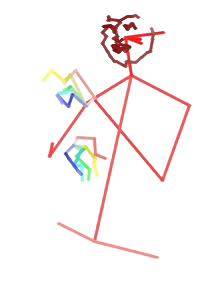} &

        \includegraphics[width=11mm, height=16mm]{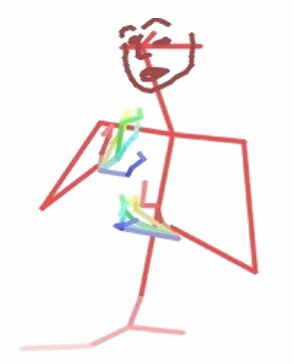} &
        
        \includegraphics[width=11mm, height=16mm]{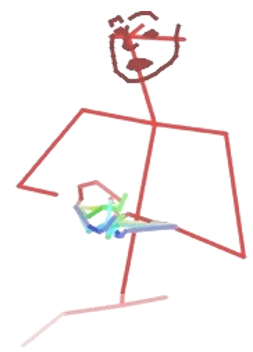}  \\

        \rotatebox{90}{\phantom{A}{\small MOMP}} &

        \includegraphics[width=14mm, height=17mm]{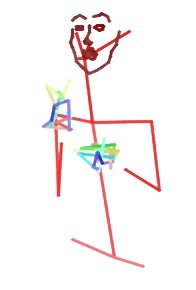} &

        \includegraphics[width=14mm, height=17mm]{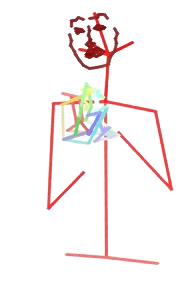} &

        \includegraphics[width=14mm, height=17mm]{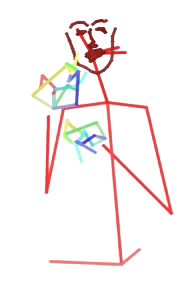} &

        \includegraphics[width=14mm, height=17mm]{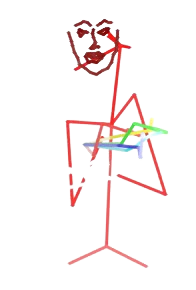} &
        
        \rotatebox{90}{\phantom{Aaa}{\small A2S}} &
        
        \includegraphics[width=14mm, height=17mm]{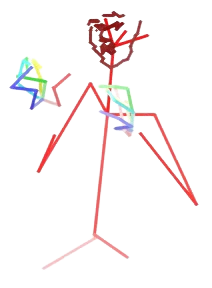} &

        \includegraphics[width=14mm, height=17mm]{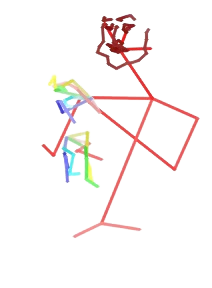} &

        \includegraphics[width=14mm, height=17mm]{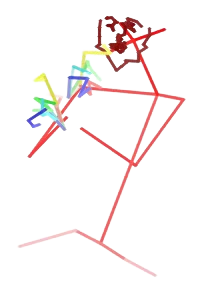} &
        
        \includegraphics[width=14mm, height=17mm]{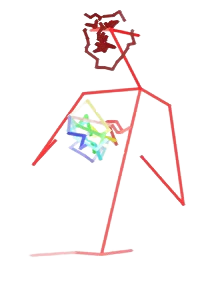}  \\

        \rotatebox{90}{\phantom{A}{\small T2M-GPT}} &

        \includegraphics[width=14mm, height=17mm]{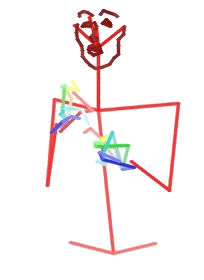} &

        \includegraphics[width=14mm, height=17mm]{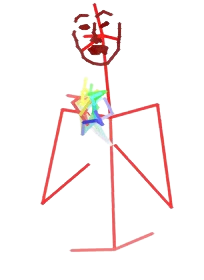} &

        \includegraphics[width=14mm, height=17mm]{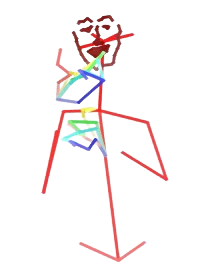} &

        \includegraphics[width=14mm, height=17mm]{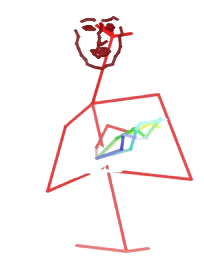} &
        
        \rotatebox{90}{\phantom{A}{\small T2A2S}} &
        
        \includegraphics[width=14mm, height=17mm]{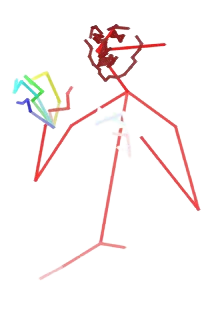} &

        \includegraphics[width=14mm, height=17mm]{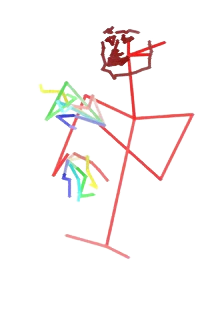} &

        \includegraphics[width=14mm, height=17mm]{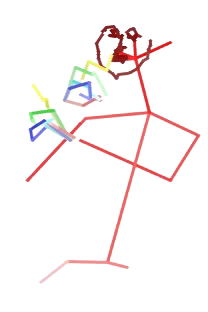} &
        
        \includegraphics[width=14mm, height=17mm]{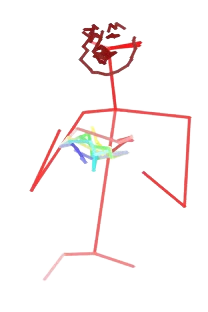}  \\

        \rotatebox{90}{\phantom{A}{\small Ham2Pose}} &

        \includegraphics[width=14mm, height=17mm]{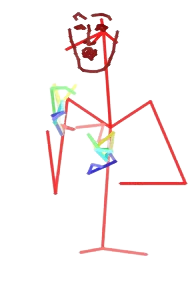} &

        \includegraphics[width=14mm, height=17mm]{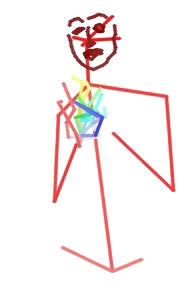} &

        \includegraphics[width=14mm, height=17mm]{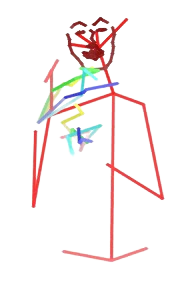} &

        \includegraphics[width=14mm, height=17mm]{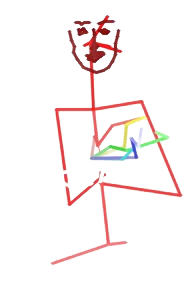} &
        
        \rotatebox{90}{\phantom{A}{\small A2T2S}} &
        
        \includegraphics[width=14mm, height=17mm]{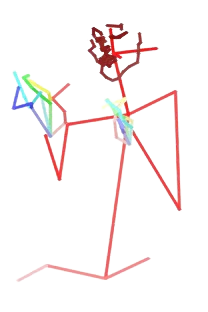} &

        \includegraphics[width=14mm, height=17mm]{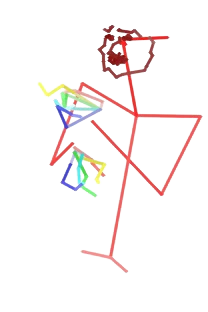} &

        \includegraphics[width=14mm, height=17mm]{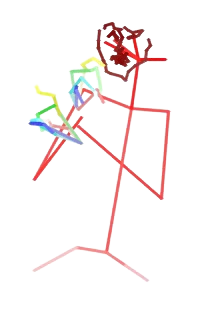} &
        
        \includegraphics[width=14mm, height=17mm]{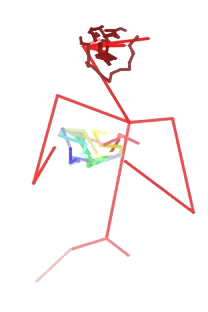}  \\

    \end{tabular}
    }
    \vspace{-10pt}
    \caption{\textbf{Results examples~(\S\ref{sec:CS}):} \textbf{Left column:} text-to-sign generation stream, \textbf{right column:} audio-to-sign generation stream. Under given conditions, our MS2SL can generate signs that are more semantically consistent with the spoken description and have more precise keypoints.}
    \label{fig:results_example}
    \vspace{-12pt}
\end{figure*} 

%% file: sections/conclusion.tex
\vspace{1pt}
\section{Conclusion}
\label{sec:con}
We explore a unified framework that combines diffusion and pretrained models to generate sign language from spoken depictions.
We surpass other competitors and solidify this classic framework as a highly competitive method for SLP.
MS2SL effectively handles diverse modalities of data for analysis and decoupling.
Despite its advancements, our model struggles with maintaining contextual flow in generation, and MS2SL cannot handle lengthy data, which is a future focus. 
Our research pioneers direct sign language generation from speech, offering some insights to advance the community. 

%% file: sections/Limitations.tex
\renewcommand\thesection{\Alph{section}}
\renewcommand*{\theHsection}{appedix.\thesection}
\setcounter{section}{0}
\setcounter{figure}{3}
\setcounter{table}{6}
\setcounter{equation}{4}
\section*{Limitations}
Despite significant advancements, our method still faces key technical limitations. First, the complexity and fluidity of authentic sign language are challenging to fully capture and reproduce, as it involves not just hand movements but also facial expressions, body language, and the speed of gestures. 
Moreover, converting text or speech into sign language involves complex natural language processing challenges, especially in handling grammar and semantics. 
Lastly, MS2SL struggles to effectively generate long sequences of key movements, limiting the coherence and completeness of sign language expression. These limitations indicate that, while the potential of sign language generation technology is immense, significant technical barriers still need to be overcome to achieve comprehensive and precise sign language communication.
These are also the directions we are committed to addressing in the future. 

%% file: sections/appendix.tex
\renewcommand\thesection{\Alph{section}}
\renewcommand*{\theHsection}{appedix.\thesection}
\setcounter{section}{0}
\setcounter{figure}{3}
\setcounter{table}{6}
\setcounter{equation}{4}

\clearpage
\vspace{0.5em}
\textbf{\centering \Large ~~~~~Supplementary Material}
\vspace{1.0em}
\setcounter{page}{1}

In the appendix, we provide the following components that offer a more comprehensive understanding of our method:
\vspace{-3pt}
\begin{itemize}[leftmargin=*]
	\setlength{\itemsep}{0pt}
	\setlength{\parsep}{-2pt}
	\setlength{\parskip}{-0pt}
	\setlength{\leftmargin}{-10pt}
	\vspace{-4pt}
  \item \S\ref{sec:AD}: More Experimental Results.
  \item \S\ref{sec:albation}: Architecture Details.
  \item \S\ref{sec:Impact}: Impacts.
\end{itemize}
\vspace{-3pt}
We employ GPT-3.5 to refine and enhance our writing. We are immensely grateful for the substantial assistance provided by GPT.
\section{More Experimental Results}
\vspace{-3pt}
\label{sec:albation}
We$_{\!}$ conducte$_{\!}$ multiple$_{\!}$ experiments$_{\!}$ and report the results on PhoenixT~\cite{DBLP:conf/cvpr/CamgozHKNB18}, and we does not conduct comparative experiments for the audio-to-sign generation stream due to the absence of any audio data in PhoenixT.
As shown in Table~\ref{tbl:PDM} and ~\ref{tbl:DTPM}, we report the diagnostic results on PhoenixT. 
After integrating the diffusion process into MS2SL, we note modest enhancements. In particular, BLEU-1 score shows a notable improvement, rising by $1.88$, and ROUGE score experience a increment of $2.11$~(Table~\ref{tbl:PDM}).
\begin{table}[h]
    \makeatletter\def\@captype{table}\captionsetup{width=1.\linewidth}
     \vspace{-4pt}
    \centering\small
    \resizebox{1.\linewidth}{!}{
        \setlength\tabcolsep{4pt}
        \renewcommand\arraystretch{1.2}
        \begin{tabular}{c||ccccc}
        \thickhline
           Methods & BLEU-4 & BLEU-3 & BLEU-2 & BLEU-1 & ROUGE \\
           \hline
            0 &  \et{10.19.}{.01} & \et{12.58}{.04} & \et{18.48}{.05} & \et{31.92}{.04} & \et{33.80}{.01} \\
           5 &  \et{10.91}{.03} & \et{13.11}{.02} & \et{21.47}{.05} & \et{33.8}{.06} & \et{35.91}{.00} \\
           10&  \etr{12.77}{.06} & \etr{15.81}{.07} & \et{22.04}{.03} & \et{36.41}{.01} & \et{36.63}{.03} \\
           15 &  \et{12.97}{.02} & \et{15.06}{.01} & \etr{22.25}{.05} & \etr{36.73}{.02} & \etr{37.10}{.02} \\
            \hline
    \end{tabular}
    }
    \vspace{-6pt}
    \caption{\textbf{Denoising steps.}} 
    \label{tbl:PDM}
    \vspace{-8pt}
\end{table}

In$_{\!}$ the$_{\!}$ comparison$_{\!}$ of$_{\!}$ pre-trained$_{\!}$ models~(Table~\ref{tbl:DTPM}), the conclusion is similar to that with How2Sign, indicating no significant differences among various text pre-training models.$_{\!}$ 
This$_{\!}$ is$_{\!}$ due$_{\!}$ to$_{\!}$ the relatively small dataset and vocabulary size of PhoenixT, for which the current models are sufficiently. 
\begin{table}[h]
    \makeatletter\def\@captype{table}\captionsetup{width=.9\linewidth}
    \vspace{-3pt}
    \centering\small
    \resizebox{1.\linewidth}{!}{
        \setlength\tabcolsep{1pt}
        \renewcommand\arraystretch{1.7}
        \begin{tabular}{c||ccccc}
        \thickhline
           Methods & BLEU-4 & BLEU-3 & BLEU-2 & BLEU-1 & ROUGE \\
           \hline
           GT &  \et{20.53}{.01} & \et{25.13}{.03} & \et{32.81}{.04} & \et{44.01}{.02} & \et{45.61}{.03} \\
           CLIP~\cite{DBLP:conf/icml/RadfordKHRGASAM21} &  \etr{12.77}{.06} & \etr{15.81}{.07} & \et{22.04}{.03} & \et{36.41}{.01} & \etr{36.63}{.03} \\
           Bert~\cite{DBLP:conf/naacl/DevlinCLT19} & \et{12.52}{.02} & \et{15.76}{.04} & \etr{22.39}{.03} & \etr{37.13}{.02} & \et{36.45}{.05} \\
            \hline
        \end{tabular}
    }
    \vspace{-6pt}
    \caption{\textbf{Different text pre-trained models.}}
    \label{tbl:DTPM}
    \vspace{-8pt}
\end{table}
\begin{table*}[h]
\centering
\resizebox{1.\linewidth}{!}{
                \setlength\tabcolsep{9pt}
                \renewcommand\arraystretch{1.4}
\begin{tabular}{@{}cccccc@{}}
\toprule
Module & Text Encoder & Audio Encoder & Sign Encoder & Step Encoder & Noise Encoder \\
 \midrule
Input      & text         & audio  &  sign keypoints & step number & sign noise       \\
Feature extraction     & CLIP~\cite{DBLP:conf/icml/RadfordKHRGASAM21}    & Hubert~\cite{DBLP:journals/taslp/HsuBTLSM21}  & Positional Encoding & nn.Embedding & Positional Encoding \\
\cline{1-6}
\multirow{2}{*}{Embedding Generation}   &   Attention Blk $\times 2$     & Attention Blk $\times 2$ & Attention Blk $\times 2$ & Attention Blk $\times 2$ & Attention Blk $\times 2$ \\ 
                                        & MLP & MLP & MLP & MLP & MLP \\
\cline{1-6}
  \multirow{2}{*}{Embedding Fusion} & \multicolumn{2}{c}{Length prediction (MLP)} & \multicolumn{3}{c}{-} \\
                                    & \multicolumn{5}{c}{Concatenation}\\
\cline{1-6}
\multirow{2}{*}{Sign Prediction }  & \multicolumn{5}{c}{Attention Blk $\times 6$} \\
                                   & \multicolumn{5}{c}{MLP}                   \\               
\bottomrule
\end{tabular}
}
\vspace{-5pt}
\caption{\textbf{Network architecture of the sign predictor~(\S\ref{sec:AD}).}}
\label{tbl:NA}
\vspace{-13pt}
\end{table*}
\vspace{-5pt}
\section{Architecture Details}
\label{sec:AD}
\vspace{-5pt}
The sign predictor, designed for predicting noise at each step~$h$ in the diffusion process~\cite{nichol2021improved,DBLP:conf/cvpr/Shalev-Arkushin23}, boasts a streamlined network architecture with several specialized modules.
Table~\ref{tbl:NA} details the parameter configurations of each module in MS2SL.

\noindent\textbf{Encoders}. We employ a total of five encoders to process different types of input content. 
Each encoder consists of two attention layers and a Multi-Layer Perceptron (MLP).
Attention mechanisms~\cite{radford2018improving} in each encoder enable the model to focus on the most relevant features of input data, enhancing its ability to extract and learn complex patterns.
MLP further processes those focused information to generate embeddings~$\bm{e}_t$,~$\bm{e}_a$,~$\bm{e}_s$,~$\bm{e}_h$~and~$\bm{e}_n$, introducing non-linear transformations to add depth to the analysis and enabling the extraction of higher-level features.

\noindent\textbf{Producer.} The producer is a central component of the model, responsible for synthesizing and outputting the final sign predictions.
MS2SL utilizes the attention mechanism to learn the relationships between different input content, gathered and processed by the encoders.
We utilize six multi-head attention blocks.
Finally, we also use an MLP to transform the predicted features into coordinates for $137$ sign keypoints in How2Sign~\cite{DBLP:conf/cvpr/DuartePVGDMTG21} and PHOENIX14T~\cite{DBLP:conf/cvpr/CamgozHKNB18}.

We also designed a length predictor to forecast the length of the generated sign language sequences. 
By accurately predicting the sequence length, the length predictor helps maintain the coherence and consistency of the model's outputs, ensuring they are accurate not only in content but also in their temporal unfolding. 
To reduce the overall parameters of the model, we employed separate predictors for estimating the length of the input text and audio, respectively.
\section{Impacts}
\label{sec:Impact}
\vspace{-5pt}
Sign language production technology has significant impacts in both social and technological areas. 
Socially, it greatly enhances accessible communication, improving information access and interaction for deaf and hard of hearing individuals, especially in daily life, education, and work environments. 
It can foster social inclusiveness, aiding in the dismantling of communication barriers and facilitating the integration of the deaf community into broader society.
SLP also serves as an educational tool, aiding deaf students in better understanding and absorbing information and facilitating the learning of sign language for hearing individuals. 
Technologically, the advancement of SLP drives progress in image recognition, natural language processing, and machine learning. 
This involves tackling challenges such as multimodal learning, text and audio comprehension, content generation, and data scarcity simultaneously.
We conduct cyclic consistency learning on a joint embedding space, providing effective insights for niche domains.
It aslo poses some potential risks, including insufficient accuracy, cultural nuances, and misinterpretations. 